\pdfoutput=1

\documentclass[11pt]{article}

\usepackage[]{emnlp2022}

\usepackage{times}
\usepackage{latexsym}

\usepackage[T1]{fontenc}

\usepackage[utf8]{inputenc}

\usepackage{microtype}
\usepackage{makecell}

\usepackage{soul}
\usepackage{url}
\usepackage[utf8]{inputenc}
\usepackage{graphicx}
\usepackage{amsmath}
\usepackage{booktabs}
\usepackage{colortbl}
\usepackage{multirow}
\usepackage{multicol}
\usepackage{setspace}  

\definecolor{ForestGreen}{rgb}{0.13, 0.55, 0.13}

\newcommand{\smal}{\fontsize{10pt}{10pt}\selectfont }

\newcommand{\eat}[1]{}
\newcommand{\red}[1]{\textcolor{red}{#1}}
\newcommand{\blue}[1]{\textcolor{blue}{#1}}

\newcommand{\todo}[1]{\textcolor{red}{[\textsc{TODO: }#1 ]}}
\newcommand{\camera}[1]{\textcolor{black}{#1}}

\newcommand{\sgap}{\hspace*{2mm}}
\newcommand{\gap}{\hspace*{1cm}}
\newcommand{\lgap}{\hspace*{1.5cm}}
\newcommand{\ra}{$\rightarrow$}

\newcommand{\gapx}{\hspace*{2mm}}
\newcommand{\gapxx}{\hspace*{4mm}}
\newcommand{\gapxxx}{\hspace*{6mm}}

\newcommand{\gapxxxxx}{\hspace*{10mm}}

\newcommand{\gapxxxxxxxx}{\hspace*{16mm}}

\newcommand{\EntailmentWriter}{Entailer}

\newcommand{\bfit}[1]{\textbf{\textit{#1}}}

\usepackage{quoting}
\newenvironment{myquote}{                   
  \parskip 0mm \begin{quoting}[vskip=0mm,leftmargin=2mm]}{
\end{quoting}}
\newenvironment{ite}{                     
     \parskip 0cm \begin{itemize} \parskip 0cm \parsep 0cm \itemsep 0cm \topsep 0cm}{
        \end{itemize}} 
\newenvironment{enu}{                   
     \parskip 0cm \begin{list}{}{\parsep 0cm \itemsep 0cm \topsep 0cm}}{
       \end{list}} 
\newenvironment{des}{                 
     \parskip 0cm \begin{list}{}{\parsep 0cm \itemsep 0cm \topsep 0cm}}{
       \end{list}} 

\newcommand{\quotebox}[1]{\begin{myquote}\fbox{\parbox{0.9\columnwidth}{#1}}\end{myquote}}

\usepackage{algorithm}
\usepackage{algorithm}
\usepackage[noend]{algpseudocode}
\usepackage{ulem}

\title{\EntailmentWriter{}: Answering Questions with Faithful and Truthful Chains of Reasoning}

\author{Oyvind Tafjord, Bhavana Dalvi Mishra, Peter Clark \\
Allen Institute for AI, Seattle, WA \\
{\texttt{\{oyvindt,bhavanad,peterc\}@allenai.org}}
}

\begin{document}
\maketitle

\begin{abstract}
Our goal is a question-answering (QA) system that can show how its answers are implied by
{\it its own internal beliefs} via a {\it systematic chain of reasoning}. Such a capability would allow better understanding of {\it why} a model produced the answer it did. Our approach is to recursively combine a trained backward-chaining model, capable of generating a set of premises entailing an
answer hypothesis, with a verifier that checks that the model itself believes those premises (and the entailment itself)
through self-querying.
To our knowledge, this is the first system to generate multistep chains that are both {\it faithful} (the answer follows from the reasoning) and {\it truthful} (the chain reflects the system's own internal beliefs). In evaluation using two different datasets, users judge that a majority (70\%+) of generated chains clearly show how an answer follows from a set of facts - substantially better than a high-performance baseline - while preserving answer accuracy. By materializing model beliefs that systematically support an answer, new opportunities arise for understanding the model's system of belief,
and diagnosing and correcting its misunderstandings when an answer is wrong.
\end{abstract}

\section{Introduction}

\begin{figure}[t]
\centering
     \includegraphics[width=\columnwidth]{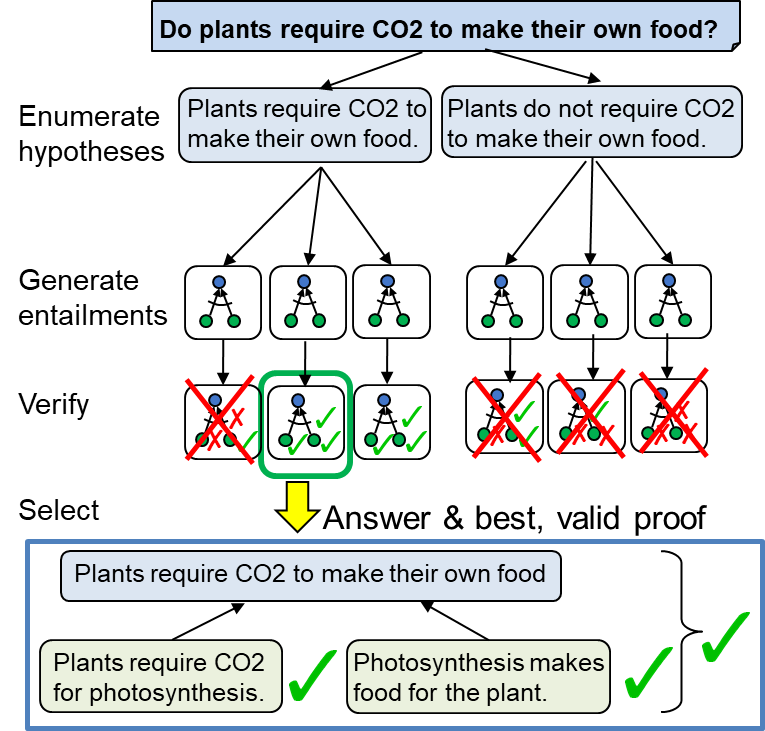}	   
\caption{Given a question, \EntailmentWriter{} searches for an answer hypothesis that is supported by an entailment proof. First it over-generates candidate proofs, then it removes those that the model itself does not “believe” (i.e., confirms via self-querying that it considers all the generated proof elements to be true). Finally it selects the best verified proof. Multistep proofs are generated by iteratively backward chaining  on the premises (Section~\ref{generating-entailment-trees}).
\label{entailmentwriter}}
\vspace{-3mm}
\end{figure}

Although pretrained language models (PTLMs) have shown remarkable question-answering (QA) performance, it is often
unclear {\it why} their answers follow from what they know. While there has been substantial work on training
models to also generate explanations for their answers \cite{wiegreffe2021teach},
or produce them via few-shot prompting, e.g., ``chains of thought'' \cite{Wei2022ChainOT},
those explanations may
not be {\it faithful} (the answer does not necessarily follow from them) and may not be {\it truthful}, in the sense that
the language model itself does not believe\footnote{
We here adopt a simple \camera{operational} definition of belief, namely that a model believes X if it answers "yes" to the question "Is X true?".
Other definitions could also be used.} the explanation statements that it generated. Rather, our
goal is to generate answers that {\it systematically follow} from the model's own internal beliefs,
materializing those beliefs as explicit statements that can then be inspected. Such a capability
offers new opportunities for understanding, diagnosing, and ultimately correcting errors
in a language model's behavior.

\begin{figure*}
\noindent\fbox{\parbox{\textwidth}{
\small{
Q: A magnet will stick to (A) a belt buckle (B) a wooden table (C) a plastic cup (D) a paper plate  \\
A: (A) A magnet will stick to a belt buckle {\bf because} \\
\gap      A belt buckle is sometimes magnetic. {\bf because} \\
\lgap        Metal is sometimes magnetic. \\
\lgap         A belt buckle is made of metal. \\
\gap      A magnet will stick to magnetic metals. \\
\\
Q: You can make a telescope with a (A) straw (B) Glass (C) Candle (D) mailing tube \\
A: (D) You can make a telescope with a mailing tube. {\bf because} \\
\gap      A telescope is made of a tube for observing / seeing. \\
\gap      A mailing tube is a kind of tube. \\
\\
Q: Quartz may produce rainbows when light is shined (A) around the crystal's area (B) through any of its sides (C) in the room its in (D) in to a mirror at it \\
A: (B) Quartz may produce rainbows when light is shined through any of its sides. {\bf because} \\
\gap     A rainbow is produced when light shines through a prism. {\bf because} \\
\lgap        The rainbow is made of all different colors in visible light. \\
\lgap        A prism can split light into different colors.  \\
\gap     A quartz is a kind of prism.
}}}
\caption{Questions (from the OBQA dataset) and \EntailmentWriter{}'s answers, showing its chain of reasoning. 
  \label{example-proofs}}
\vspace{-2mm}
\end{figure*}

Our approach uses a combination of generation and verification,
implemented in a system called \EntailmentWriter{}\footnote{\camera{\EntailmentWriter{} data and models are available at  \url{https://allenai.org/data/entailer}}}.
Chains are constructed by backward chaining from candidate answers,
recursively using a language model (LM) trained for a single backward-chaining step.
For each step, \EntailmentWriter{} over-generates candidate entailments, then
filters out those that do not conform to its own internal knowledge (``beliefs'')
by self-querying, asking itself whether (a) the generated premises (leaves
of the proof step) are true, and (b) each entailment step is valid (Figure~\ref{entailmentwriter}).
It then recursively backward-chains on premises until the overall proof
confidence cannot be further improved (or a depth limit $d$ is reached).
Finally, the candidate answer supported by the highest-scoring chain of
a reasoning is returned. As a result, the system has materialized some of 
its latent knowledge from which the selected answer follows.
Most significantly, the resulting proof is thus both {\it faithful} (the answer
follows from the proof) and {\it truthful} (the proof reflects the system's beliefs),
providing a previously unavailable window into the model's beliefs about the
world and their implications, e.g., Figure~\ref{example-proofs}.

To train the \EntailmentWriter{} model, we use a
combination of the existing EntailmentBank dataset \cite{entailmentbank},
plus a new crowd-annotated dataset that we construct by bootstrapping
(train an initial model, generate candidate entailment examples
with it, then annotate those examples as extra training data). The model is 
then frozen, and \EntailmentWriter{} is then applied {\bf zero-shot} to new
datasets, i.e., \EntailmentWriter{} is a treated as a general-purpose,
fixed model specialized for reasoning, rather than requiring fine-tuning
for new tasks.

We evaluate \EntailmentWriter{} on two existing datasets, OBQA \cite{obqa} and QuaRTz \cite{quartz}.
We find that its reasoning-based QA accuracy is similar
to its direct QA accuracy, with the advantage that a supporting chain of
reasoning is also produced.  We also perform a human evaluation, and
find that 70\% of time users judge the chains to clearly show how an 
answer followed from their premises, substantially higher than for
explanations produced by a comparable high-performance QA system, Macaw \cite{macaw}.
Our contributions are thus:
\begin{enu}
\item[1.] The first system to generate chains of reasoning showing how
answers are systematically implied by a {\it model's own internal beliefs},
making relevant model beliefs explicit. The chains are both
\bfit{faithful} (the answer follows from the reasoning) and \bfit{truthful}
(the chain reflects the system's own beliefs).
\item[2.] A new, crowdsourced dataset of multi-premise entailments,
doubling the amount of data available in EntailmentBank \cite{entailmentbank},
and including examples of both positive and negative entailments
(EntailmentBank only includes positive examples)\footnote{The dataset is provided in the supplementary material.}.
\end{enu}

\section{Related Work \label{related-work}}

\noindent
{\bf Systematic Reasoning:}
Several recent systems have demonstrated the ability to perform {\it systematic}
reasoning directly over natural language (Natural Language Inference \cite{Manning2009NaturalLI}),
namely deriving conclusions from known facts via step-wise application of well-defined
inference operations. One approach is to retrain a black-box model end-to-end \cite{ruletaker},
but has been limited to small rulebases. An alternative approach, which we follow here, is to have an
outside loop around a model, where the model generates individual inference steps (i.e., rules), and a
controller chains them together. SCSearch \cite{scsearch}, NLProofS \cite{NLProofS}, IRGR \cite{IRGR}, ProofWriter (iterative version) \cite{proofwriter}, \camera{and Selection-Inference \cite{Creswell2022SelectionInferenceEL}}
do this in a forward-chaining fashion, MetGen \cite{metgen} does this bidirectionally, 
while Braid \cite{braid} (like us) does this in a backward-chaining fashion.
In all these systems, the required facts were expected to be provided explicitly to the model.
In contrast, \EntailmentWriter{}'s reasoning uses its own internal, latent knowledge,
as well as (optionally) externally provided facts.
LeapOfThought \cite{leapofthought} demonstrated that reasoning with a combination of
implicit and explicit knowledge was possible for simple 1-step inferences.
We expand this for multi-step inference, and (unlike LeapOfThought) have the system
also explicitly articulate the implicit knowledge it uses, and its chain of reasoning.

Recent work has shown that generating a free-form explanation (``chain of thought'')
before an answer can also improve performance on a variety of
tasks \cite{Wei2022ChainOT,Cobbe2021TrainingVT,Nye2021ShowYW}.
In these works, however, the explanations are unstructured,
and there are no claims of faithfulness that the answer follows from the
generation, nor that the explanations themselves represent model beliefs.

\noindent
{\bf Materializing a Model's Internal Knowledge:} Pretrained LMs 
contain a vast amount of knowledge, and can be thought of as a
kind of knowledge base to tap into \cite{Petroni2019LanguageMA}.
Recent work has shown that this latent knowledge can be materialized 
as explicit English sentences or a knowledge graph using generative techniques,
e.g., COMeT \cite{Bosselut2019COMETCT}, ParaCOMET \cite{Gabriel2021ParagraphLevelCT}.
Our work with \EntailmentWriter{} similarly materializes its latent knowledge, but here
in a {\it goal-directed} way, namely by producing a faithful chain of
reasoning from facts it validates (``believes'') as true to an answer.
This articulation can be seen as a kind of self-talk,
where a self-generated context can improve QA \cite{self-talk}. However,
here our generations are not used as context for opaque problem-solving,
but are assembled into a well-defined chain of reasoning.

\noindent
\camera{
{\bf Beliefs:} We refer to the model's factual opinions as ``beliefs'' rather
than ``knowledge'' because those opinions may be wrong. In general,
an agent can be said to believe p if it acts as if p was true \cite{sep-belief}.
Following \cite{beliefbank}, we take a simple, syntactic operationalization 
of this, namely the agent answers ``yes'' to the question ``p?'',
but also note that more semantic versions could be used, e.g.,
the agent also answers ``yes'' to paraphrases and implications of p.
In general, models can sometimes be inconsistent in their beliefs
\cite{Elazar2021MeasuringAI,Kassner2020NegatedAM,Ribeiro2019AreRR}.
For our purposes here, we simply note that such inconsistencies may occasionally
exist, and that techniques for inconsistency resolution could be applied in
future to reduce these, e.g., \cite{beliefbank,Li2019ALF}.
}

\eat{
\noindent
{\bf Overgenerate-and-Filter:} Recent work has shown that an overgenerate-and-filter
approach can be particularly effective for high-quality generations, e.g.,
\cite{Cobbe2021TrainingVT}, AlphaCode \cite{alphacode}. We follow a similar
approach here to inference chain generation.

\todo{Something about the recent "chain of thought" reasoning in few-shot models? (https://arxiv.org/pdf/2201.11903.pdf, https://arxiv.org/pdf/2203.11171v1.pdf)}
}

\begin{figure}
\centering
     \includegraphics[width=\columnwidth]{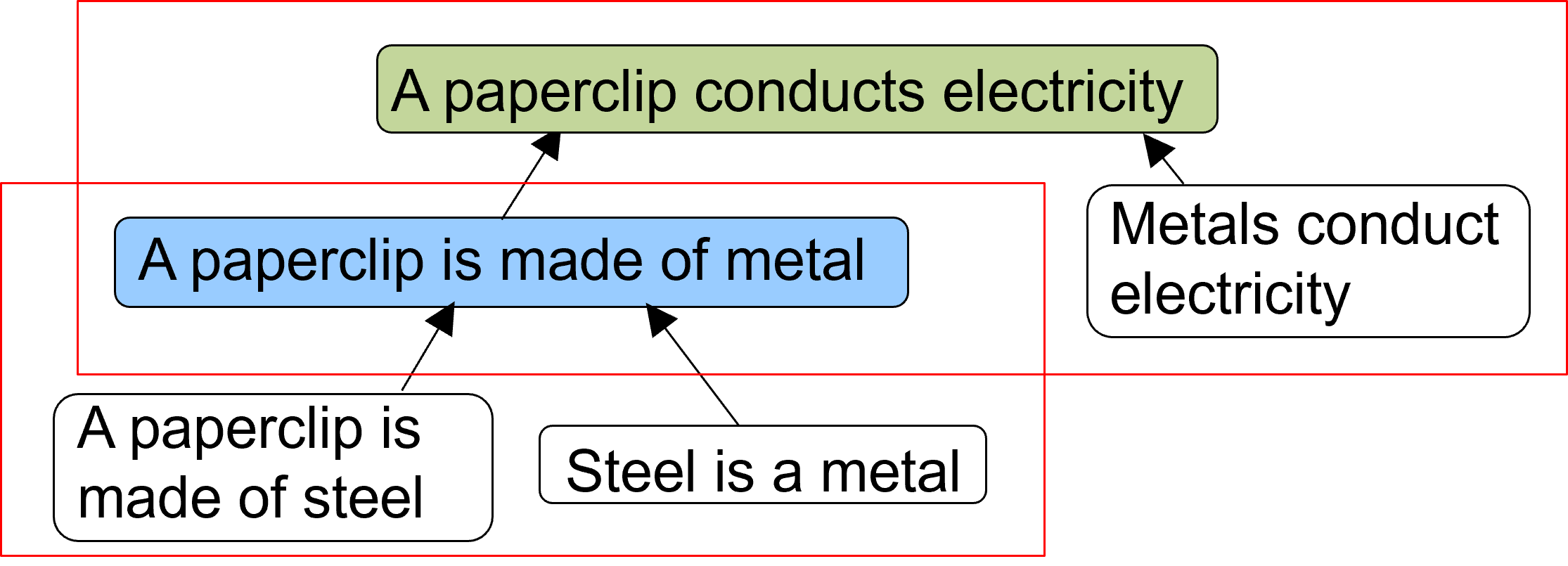}	   
\caption{An {\it entailment tree} is a set of multi-premise, 1-step entailments
(red boxes) showing how the hypothesis (root node, green) is entailed from the leaf nodes (white).
If all the leaf nodes are true, and all the 1-step entailment relations are valid,
then we say the tree is a {\it valid chain of reasoning} for the hypothesis. \label{example-entailment}}
\vspace{-3mm}
\end{figure}

\section{Approach}


Like several previous systems (Section~\ref{related-work}), \EntailmentWriter{} treats reasoning
as Natural Language Inference (NLI). In NLI, the basic unit of knowledge is (represented as)
a sentence rather than a structure, and a proof\footnote{
We use the word ``proof'' for convenience but note that the term is somewhat approximate,
as entailment ``proofs'' do not have the guarantees of formal, deductive proofs.} is
a tree of multi-step, multi-premise entailments, e.g., Figures~\ref{example-proofs} and ~\ref{example-entailment}.

Within this framework, given a question, \EntailmentWriter{} first generates
candidate answers, then tries to prove each one, selecting the answer
with the highest-scoring proof.{\footnotesize{\ }}We now describe these steps. 

\eat{
Within this framework, \EntailmentWriter{}'s novel contribution is to construct
entailment trees grounded in the system's latent ``beliefs'' about the world, rather
than (just) external knowledge sources. Following \cite{beliefbank}, we say
a system believes a fact $F$ if it answers ``yes'' to the question ``Is $F$ true?'' - 
in other words, we establish (a simple notion of) belief through self-querying.
Note we use the term ``belief'' rather than ``knowledge'', as the system's beliefs
may be wrong. Given a hypothesis $H$, \EntailmentWriter{} searches for a chain
of reasoning showing how $H$ follows from facts it believes. Similarly, given multiple
competing hypotheses (e.g., from a multiple-choice question) \EntailmentWriter{}
searches for the one supported by a chain of reasoning (or the one supported
by the best chain, if more than one).
}


\begin{table*}
\noindent\fbox{\parbox{\textwidth}{
{\small
\setlength{\tabcolsep}{2pt}	
\begin{tabular}{ll} 
{\bf Angle} &  	{\bf Input $\rightarrow$ Output} (example) \\
$H \rightarrow P$  & "H: A paperclip is made of metal. P:" $\rightarrow$ "[PREMISE] A paperclip is made of steel. [PREMISE] Steel is a metal." \\
$H \rightarrow S_d$ & "H: A paperclip is made of steel. V:" $\rightarrow$ 0.995 \\
$PH \rightarrow S_e$ & "H: A paperclip is made of metal. P: [PREMISE] A paperclip is made of steel. [PREMISE] Steel is a metal. I:" \\
 & \hspace*{2cm} $\rightarrow$ 0.998
\end{tabular}
}}} 
\caption{Examples of the three input/output angles used by \EntailmentWriter{}. The first generates a candidate entailment rule P$\vdash$H given H. The second and third score whether each premise, and the entailment itself, is valid, using tokens V/I in the input to indicate that $S_d$/$S_e$ is the desired output. \label{angles}}
\end{table*}

\begin{figure*}
      \includegraphics[width=\textwidth]{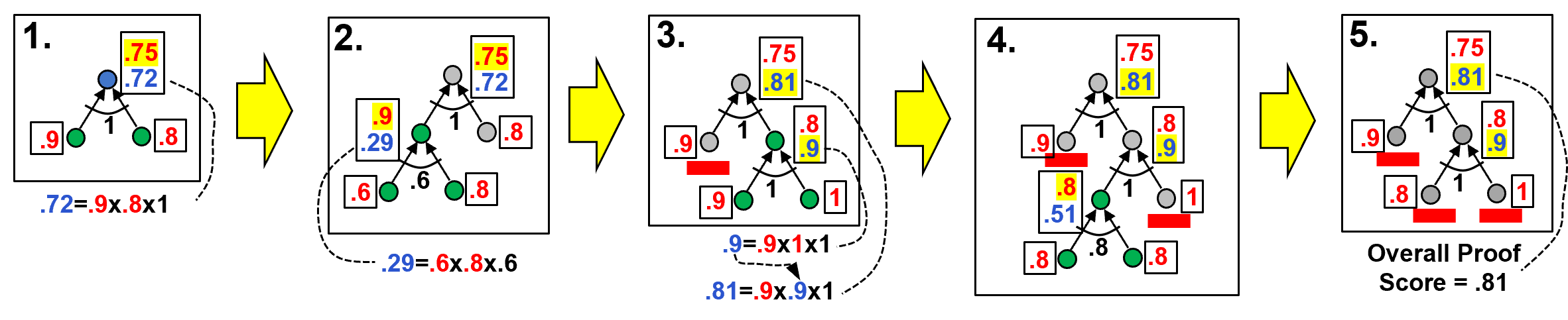}	   
     \caption{The entailment tree is grown recursively, the algorithm looking for the best tree (maximizes the overall score of the root node). Each node has a fixed, direct (“fast”) score (in red), and (for internal nodes) a proof (“slow”) score (in blue) computed from its children. The overall node score (highlighted) is the highest of the two.
     If expanding a node increases its overall score (e.g., step 3), that increase is propagated upwards and recursion continues. If expansions cannot improve a node’s score further (e.g., steps 2 and 4), the expansions are pruned and that node becomes a leaf (red bars). \label{growing-a-tree}}
     \vspace{-2mm}
\end{figure*}

\subsection{Hypothesis Generation \label{qa2d}}

Given a question, \EntailmentWriter{} first generates candidate answers and converts
these into declarative hypotheses (e.g., ``Is the sky (A) blue (B) yellow'' $\rightarrow$
\{ $H_1$ = ``The sky is blue.'', $H_2$ = ``The sky is yellow.'').\footnote{Conversion of a QA pair to a declarative hypothesis D uses a custom T5-11B model trained on the QA2D dataset \cite{qa2d}.}
An $N$-way multiple choice question yields $N$ hypotheses.
A true/false question yields 2 hypotheses. For open-ended questions, \EntailmentWriter{}
first collects $N$ candidate answers generated by an external source (Macaw \cite{macaw}
using nucleus sampling \cite{nucleus-sampling}) then forms $N$ hypotheses from them.

\eat{
\subsection{Entailment Semantics}

A chain of reasoning is a multi-step entailment tree containing a hierarchy of 1-step, multi-premise entailments from two or
more premises $P = \{p_{1},...,p_{n}\}$ to a conclusion (either a higher-level
premise or the root hypothesis $H$). For each 1-step entailment,
we say the root node is true if all its children are also true {\it and} the entailment
relation between the children and root is valid.\footnote{
A valid inference is informally defined as one that ``a person would typically infer.'' \cite{Dagan2013RecognizingTE}.
We operationalize this by training a model on user-annotated data (Section~\ref{training}).}
We say the entire chain of reasoning is valid if all the leaves of the tree
are true, and all the individual 1-step entailment relations are also valid.
If the reasoning is valid, then the hypothesis $H$ necessarily follows.
Conversely, if the hypothesis $H$ is false, then at least one fact in
the tree must be false and/or one of the 1-step entailment relations is invalid.
}

\subsection{Generating Entailment Trees \label{generating-entailment-trees}}

\subsubsection{Generating a Backward-Chaining Step \label{1step}}

\paragraph{Models:} \ \\
The core of \EntailmentWriter{} is generating and validating a single entailment step that entails a hypothesis. 
We define the following data types:
\begin{des}
\item[{\bf H:}] A hypothesis (English statement) to prove.
\item[{\bf P:}] A set of premises \{$p_1$,...,$p_i$\} (sentences) that together may {\it entail} the hypothesis H.
		Together, P and H form a one-deep {\it entailment step}, denoted by P $\vdash$ H.
\item[{\bf Q:}] A question posed to \EntailmentWriter{}.
\item[{\bf A:}] A candidate answer for consideration.
\item[{\bf C:}] An optional context (set of sentences) relevant to the problem. This allows \EntailmentWriter{} to
	also use external knowledge, if available, when generating a tree.
\end{des}

\noindent
We train a model (details in Section~\ref{training}) with the three input/output behaviors below (optional inputs shown in parentheses):
\begin{des}
\item[{\bf $(QAC)H\rightarrow P$}:] Given a hypothesis H, generate a set of premises P that may entail H
\item[{\bf $(QAC)H\rightarrow S_{d}$}:] Score whether the model believes that a hypothesis H (or premise $p_i$) is
true ($S_d > 0.5$) or not, (i.e. perform yes/no QA). We call this the {\bf direct} score (range 0-1).
\item[{\bf $(QAC)PH\rightarrow S_{e}$}:] Score whether the model believes
a candidate entailment (P $\vdash$ H) is valid $(S_e > 0.5$) or not, i.e., P validly entails H (range 0-1).
\end{des}
Examples of these three angles are in Table~\ref{angles}.

\paragraph{Algorithm:} \ \\
To generate a single backward-chaining step 
we adopt an overgenerate-and-filter approach, also found useful elsewhere \cite{NLProofS,Cobbe2021TrainingVT,alphacode}.
First, given $H$, we use the angle $H \rightarrow P$ to generate a set of premises $P$ that may entail $H$.
We then check that the model believes all the premises $p_i \in P$ using the $H (=p_i) \rightarrow S_{d}$ angle,
and that it also
believes the inference step $P \vdash H$ itself is valid (independent of whether the $p_i$ are true or not) using the $PH \rightarrow S_{e}$ angle.
The proof score, denoting how well the 1-step proof supports the hypothesis, is the product of the premises' and entailment scores:
\begin{myquote}
\vspace{1mm}
\sgap $s_{r\textnormal{-}1deep}(H) = (\Pi_{i} s_d(p_i)).s_e(P\vdash{}H)$
\end{myquote}
\vspace{1mm}
We repeat this $k$ times using nucleus sampling to obtain a diversity of alternative proof steps, and then select the highest-scoring one $P\vdash H$,
as illustrated in Figure~\ref{entailmentwriter}.

\subsubsection{Backward Chaining}

This one-step backward chainer is embedded in a larger control algorithm that grows a multi-step entailment tree,
searching for the best tree. 
This algorithm is illustrated in Figure~\ref{growing-a-tree} and described below. The full algorithm is in Appendix~A.

Each node $N$ in the tree has two associated scores, the direct (``fast'') score $s_{d}$, denoting the model's direct belief in $N$ (in red in Figure~\ref{growing-a-tree}), and (for internal nodes) the proof (``slow'') score $s_{r}$, denoting how confidently the model can {\it prove} $N$
(in blue), computed from its children. The overall score $s$ is the max of the two. The proof score $s_{r}$ is recursively defined as the product of its children’s overall scores times the entailment score:
\begin{myquote}
\vspace{1mm}
\sgap $s_{r}(N) = (\Pi_{i} s(p_i)).s_e(P\vdash{}N)$
\end{myquote}
The algorithm starts with an initial hypothesis node $H$, then
iteratively expands each leaf node $N$, looking for a proof that scores higher than the direct score $s_d$ of that node.
In other words, can the system prove $N$ with a more confidence than simply ``guessing'' if $N$ is true?
If it can, the node's overall score $s$ (max of the two) will increase, that increase is be propagated up the tree, and the
expansion is retained, e.g., step 3 in Figure~\ref{growing-a-tree}. If expansions cannot improve a node’s score further (e.g., steps 2 and 4),
the expansions are pruned and that node becomes a leaf (red bars in Figure~\ref{growing-a-tree}).

Note that because premises are self-generated rather than externally provided, this stopping condition has a different
semantics to earlier work, e.g., \cite{braid,bostrom}: Rather than
stopping when externally known facts are reached, \EntailmentWriter{} stops when
strongly believed facts are reached, and more backward chaining 
can no longer improve a hypothesis' confidence.

This whole procedure is repeated for each candidate answer hypothesis (Section~\ref{qa2d}).
Finally the system selects the answer corresponding to the hypothesis with the top-scoring proof $s(H)$.

\section{Model Training \label{training}}

The core of \EntailmentWriter{} is the model for one-step inference (Section~\ref{1step}),
with the three functionalities listed in Table~\ref{angles}.
As \EntailmentWriter{} is intended to be a general-purpose system, the model is trained one-time
for these three functionalities, and then frozen. It is then applied zero-shot to new datasets,
e.g., the evaluation in Section~\ref{experiments}.

\subsection{Data Sources \label{data-sources}}

\subsubsection{EntailmentBank} 

To train \EntailmentWriter{}'s model, we leverage (the training partition of) the EntailmentBank dataset \cite{entailmentbank}.
EntailmentBank contains 1840 multiple-choice science questions (1313 in the training partition) along with their
correct answer, a hypothesis expressing the question and answer in declarative form, and a
multistep entailment tree expressing how the hypothesis follows from a set of facts drawn from
a corpus of science facts (the WorldTree corpus \cite{worldtree}). Using the notation introduced earlier,
each EntailmentBank example is of the form:
\begin{myquote}
\hspace*{1cm} $<Q,A,H_0,\{~(P_{i} \vdash H_{i})*~\}>$
\end{myquote}
where ($P_{i} \vdash H_{i}$)* denotes a {\it set} of entailment steps
forming a tree (with root $H_{i}=H_0$), describing how the corpus facts 
entail the hypothesis $H_{0}$.
\eat{
\begin{figure}
\centering
     \includegraphics[width=1\columnwidth]{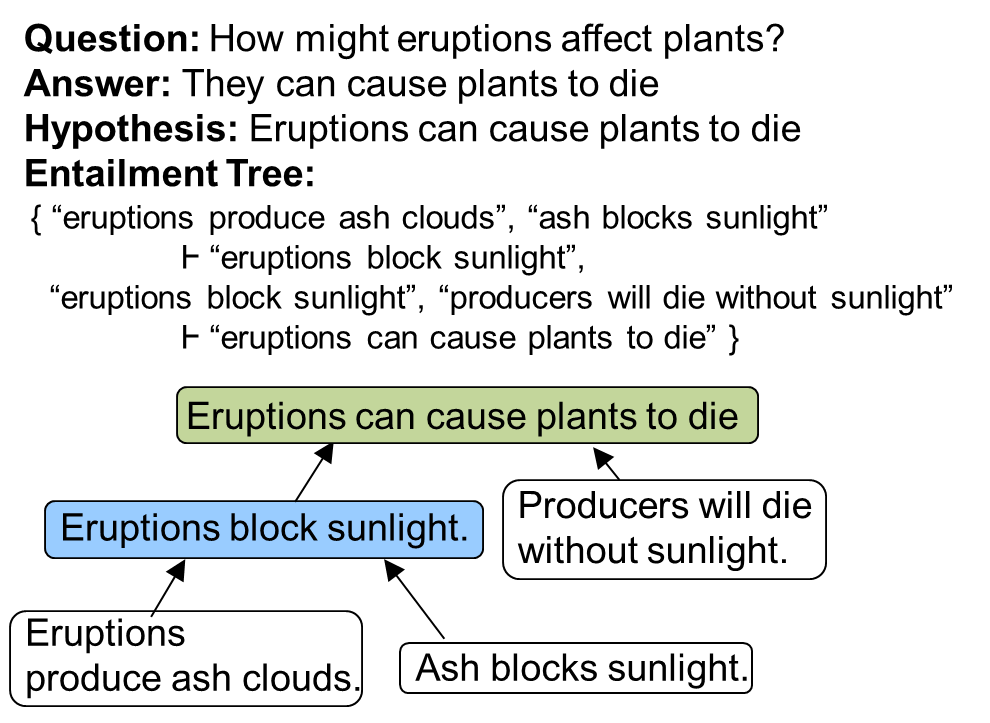}	   
\caption{An example from the EntailmentBank dataset.
\label{entailment-bank-example}}
\vspace{-8mm}
\end{figure}
}

\subsubsection{Crowdsourced Data \label{crowdsourced-data}}

EntailmentBank only contains positive examples of entailments. To obtain negative examples,
we first ran an earlier, positive-only-trained version of \EntailmentWriter{}
to generate 1-step proofs of (hypotheses for) {\it incorrect}  answer options in EntailmentBank's questions (4-way multiple choice),
resulting in (necessarily bad) proofs for the three false hypotheses.
(This was done using just the H\ra{}P angle, without verification). Note that \EntailmentWriter{}
will generate some kind of proof even for false hypotheses, e.g.,
\vspace{1mm}
\quotebox{
A rabbit has six legs {\bf because:} \\
1. A rabbit is an animal \\
2. Animals have six legs}
\vspace{1mm}
These invalid proofs will be incorrect either because a generated fact is false, and/or
because the inference itself is not a valid entailment. To distinguish these, we
use crowdworkers to assess whether the generated facts were true, and if the entailment
itself was valid. The 1313 questions result in 3939 proofs for false hypotheses.
Dropping the few with more than two premises (to simplify the crowdsourcing interface),
crowdworkers annotated 3673 proofs, using labels T/F/? for each premise and T/F/? for the entailment itself.
Each proof was annotated by 3 crowdworkers, then an additional 3 workers provided additional annotations
for cases with no initial majority verdict. Dropping premises/entailments without a final majority verdict,
we end up with 7013 additional
labeled premises
for the $H \rightarrow S_d$ angle,
and 3391 additional
labeled entailments
for the $PH \rightarrow S_e$ angle.
The crowdworker interface is in Appendix~B.

\vspace{-1mm}
\subsubsection{Optional Fields \label{optional-fields}}

We also augment the training data with duplicate examples but with additional, optional input fields:
\begin{des}
\item[{\bf QA:}] The input QA question/answer-option pair, as well as the hypothesis H
\item[{\bf C:}] A {\it context} consisting of up to 5 relevant sentences, allowing explicit external knowledge to be
           provided if available.
\end{des}
\vspace{-1mm}
To generate examples of C during training, we use a mixture of sentences drawn from (a) the gold (target)
entailment (i.e., the gold premises), and (b) sentences retrieved from a corpus of similar-styled knowledge (namely all
the leaf sentences used in the EntailmentBank entailment trees),
mixed in different ratios so the model is exposed to a mixture of high and medium relevance sentences.

\camera{In this work we do not use any context C at test time, but it is utilized in concurrent work for providing feedback to the overall system \cite{teachme}.}

Further details of training are given in Appendix~C1.

\subsection{Model Details \label{model-details}}

We train a T5-11B multi-angle model, \EntailmentWriter{}, following the multi-task setup similar to Macaw \cite{macaw}
for the three functionalities described in Table~\ref{angles}.\footnote{The scores use the token probability of generated "true" or "false" output text.}
Details of hyperparameters and other settings are provided in Appendix~C2.

\eat{
We used the default hyperparameters (including the Adafactor optimizer) in the T5 library,
fine-tune the model for 20K steps with batch size of 8, selecting the checkpoint with highest validation score (usually the final step).

At inference time, \EntailmentWriter{}'s primary function is to generate candidate entailments for each answer choice for a given question, which are then scored by the same model using its two verification angles. To generate multiple explanations for a single input (answer hypothesis, IR retrieved context), in addition to the greedy decoding we 
use nucleus sampling (for the experiments in this paper we set ``temperature= 2.0'', and ``top\_p= 0.95'').
}

\eat{
\red{Say something about C and the high/med/low aspects}, e.g.,
As discussed in Section \ref{sec:model_memory}, \EntailmentWriter{} keeps a memory of feedback received on training questions to help it assist with relevant facts at test time provided as ``context''. To be able to use such optional context at test time, it should see similar examples during training. 
Hence, during initial training we added variants of training data with context for each basic capability (H $\rightarrow$ P, $p_{i}$ $\rightarrow$ V and PH $\rightarrow$ I) of the model. 
To distinguish the context sentences by their source (user feedback vs web search) and relevance for a given question (scores produced by IR model), we divide them with [HIGH], [MEDIUM], [LOW] separators with some added noise to make the model robust. E.g. gold science fact for a question and top ranked retrievals from a textbook corpus are most likely be part of [HIGH] context, likely relevant science facts are part of [MEDIUM] context, while potentially likely facts retrieved from the Web are added to [LOW] context field. 
Finally say something about the forced prefix F training.
}

\eat{
\subsection{Model Details}
We train a T5-11B multi-angle model, \EntailmentWriter{}, following the multi-task setup similar to \cite{macaw} for both entailment generation and verification using the training data described in Section \ref{data-sources}.
We used the default hyperparameters (including the Adafactor optimizer) in the T5 library,
fine-tune the model for 20K steps with batch size of 8, selecting the checkpoint with highest validation score (usually the final step).
}

\section{Evaluation \label{experiments}}

Our goal is to generate answers supported by faithful, truthful chains of reasoning,
without significantly impacting performance. Our two corresponding reserach
questions are:
\begin{enu}
\item[1.] How does \EntailmentWriter{}'s proof-based answer accuracy compare with
          the direct QA accuracy (both zero-shot)?
\item[2.] How good are the generated entailment-based proofs? And are they
	better than those produced by a purely generative model?
\end{enu}
For the first question, we evaluate using two existing multiple-choice datasets,
namely OBQA \cite{obqa} and QuaRTz \cite{quartz}. These datasets contain
multiple-choice questions that (typically) require multihop reasoning, rather
than being simple lookup questions. 
We use the test partitions\footnote{
For QuaRTz, for synchronization with \camera{other work \cite{teachme}}, we use a
portion of the training set as our test partition here. As we are applying
Entailer zero-shot, this makes no material difference experimentally.}
with sizes 500 questions (OBQA) and 557 questions (QuaRTz).

For the second question, we collect human judgements of whether the hypothesis
clearly follows from facts in the proof, and compare its proofs with explanations
generated by a high-quality baseline model, Macaw \cite{macaw}.



\begin{figure}
\centering
     \includegraphics[width=0.9\columnwidth]{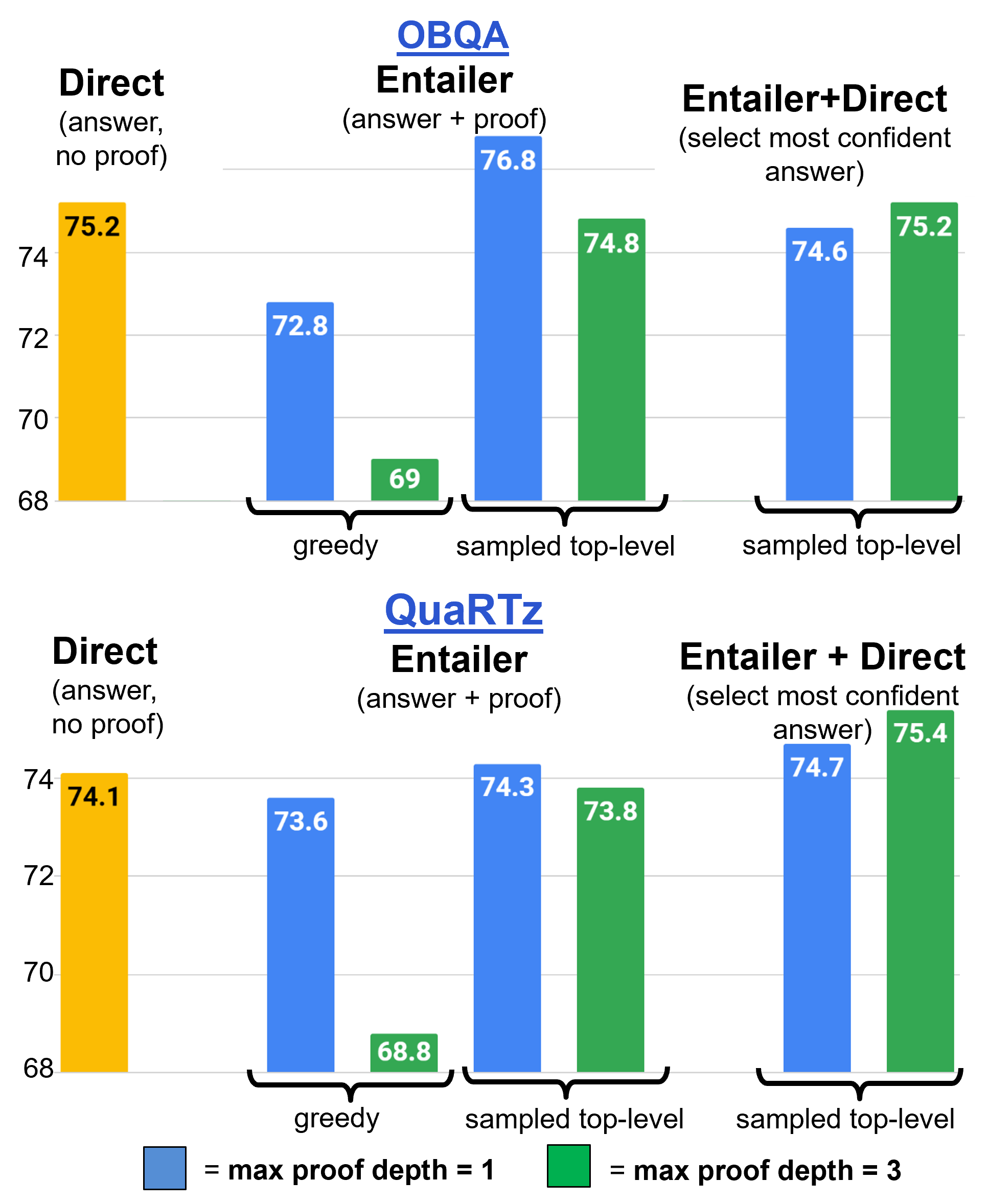}	   
\vspace{-2mm}
\caption{QA accuracy of Direct QA, \EntailmentWriter{}, and the two combined on two datasets. \label{results}}
\vspace{-3mm}
\end{figure}

\subsection{QA Accuracy}

We evaluate \EntailmentWriter{}'s proof-based QA accuracies under various conditions,
and compare them against its direct QA accuracy:

\noindent
{\bf 1. Direct QA:} Here we measure the model's ability to directly
answer the test questions (zero shot) using the $H \rightarrow S_d$ angle.
One can think of this as the {\bf ``fast thinking''} answer.
Note that this capability of the frozen model was trained on
the same data as the rest of \EntailmentWriter{}, so is a fair comparison. \\
{\bf 2. \EntailmentWriter{}:} Here we measure \EntailmentWriter{}'s ability
to answer the test questions (again zero shot) by generating, scoring, and comparing
entailment proofs for each answer option. One can think of this as the {\bf ``slow thinking''}
answer. We vary: \vspace{-1mm}
\begin{ite}
\item {\bf maximum proof depth} $d$ = 1 (blue in Figure~\ref{results}) or 3 (green)
\item {\bf degree of search:} (a) {\bf greedy:} use the first ($k=1$) generated entailment at each step, or
(b) {\bf sampled top level:} pick the best of $k=6$ entailments for expanding the root hypothesis
node. ($\approx$ six times more computationally expensive than (a)).
\end{ite} \vspace{-1mm}
Note that these proofs are are {\it faithful} explanations of why an answer was chosen,
as the selected answer by definition is always the one with the highest-scoring proof. \\
{\bf 3. \EntailmentWriter{} + Direct:} Here we combine the two by selecting the overall most
confident answer of Direct and \EntailmentWriter{} (using $c_d(H)$ or $c_r(H)$, Appendix~A).
Here, the proofs are not always faithful explanations of an answer, as
the chosen answer may not be the one with the highest-scoring proof.


\subsection{Results}

The results are shown in Figure~\ref{results}, and suggest that:

\noindent
{\bf Proof Scores are competitive with direct answer scores.} In \EntailmentWriter{}'s best configuration (sampled top-level $k=6$,
max depth $d=3$, last bar), its reason-based answers have an accuracy of 75.2/75.4 for OBQA/QuaRTz respectively, not significantly
different to the Direct scores of 75.2/74.1. This is important, as it suggests there is no significant accuracy penalty for
proof-based answers. \vspace{1mm} \\
{\bf Sampling proofs helps:} Using sampling for the top-level proofs (last 4 bars) 
always outperforms greedy proof selection by a small amount. \vspace{1mm} \\
{\bf Allowing deeper proofs does not significantly affect accuracy:} Although deeper
proofs\footnote{The distribution of proofs with different depths 1, 2, and 3 was 162, 232, and 106 respectively for OBQA,
and 123, 256, and 178 respectively for QuaRTz.}
slightly help in the combination of Entailer+Direct, and may provide more
information to a user, the accuracy differences are not significant.


When Entailer + Direct are combined, by selecting the most confident answer (last two columns),
we lose the guarantee of faithfulness, as the selected answer may not be the one with the highest-scoring
proof. In practice, this occurs for 16.8\% of the questions (OBQA), 14.2\% (QuaRTz).
In addition, we find this combination does not have significant performance gains, so has no obvious advantage in these experiments.

\eat{\camera{The absolute accuracy scores of Entailer is not a focus of this work. For comparison, the original GPT-3 (175 billion parameters) work \cite{GPT3} reported a zero-shot score on OBQA of 57.6\% (65.4\% in few-shot mode), while more recent work using rationale-augmented prompt ensembling \cite{Wang2022RationaleAugmentedEI} has reached few-shot (with rationales) scores of 91.0\% (PaLM-540B) and 88.4\% (GPT3-175B).
}}

\camera{
Note that we are measuring {\it zero-shot} performance on our test datasets, so our results are not comparable with leaderboard entries.\footnote{For a comparable datapoint on OBQA, \cite{GPT3} report zero-shot GPT3 scores of 57.6\% and few-shot 65.4\% on OBQA, so Entailer's $\sim$75\% scores are a strong zero-shot baseline for an 11B model. More recent work using rationale-augmented prompt ensembling with much larger models \cite{Wang2022RationaleAugmentedEI} has reached few-shot (with rationales) scores of 91.0\% (PaLM-540B) and 88.4\% (GPT3-175B) on OBQA.}
More importantly, though, our goal is not a state-of-the-art zero-shot model, but rather a model that can show how answers {\it systematically follow from its own internal beliefs}. Our results suggest this is possible with high reliability, and, as an additional bonus, without loss of zero-shot accuracy. Thus, rather than just answers, we now get answers supported by faithful chains of reasoning.}

\subsection{Human Judgements \label{human-judgements}}

\eat{
For our second question of evaluating the quality of \EntailmentWriter{}'s proofs, six annotators compared 1-deep \EntailmentWriter{}
proofs with explanations generated by Macaw, a state-of-the-art QA system with explanation capability \cite{macaw}.
Examples of explanations from both systems are in Appendix~D. 
Annotators scored each explanation along four dimensions, and then compared the two:}

For our second question of evaluating the quality of \EntailmentWriter{}'s proofs, we compare against
explanations generated by Macaw,\footnote{
Note that comparing with a baseline explanation generator trained on our data, i.e., just using
$H \rightarrow P$ angle without verification, would not be informative, as such explanations
would necessarily be worse: By definition, the verifiers remove explanations with either (believed to be)
false premises and/or bad inferences, hence removing the verifiers will increase the frequency of
false premises and bad inferences (confirmed through sample analysis).
Hence we instead use a strong, external system as a comparative reference point.}
a public, state-of-the-art QA system with explanation capability \cite{macaw}.
Examples of explanations from both systems are in Appendix~D.
Six annotators compared 1-deep \EntailmentWriter{} proofs with explanations from Macaw,
scoring each along four dimensions, and then comparing the two:
\begin{des}
\item[1.] Does the conclusion clearly follow from the premises?
\item[2.] Are all the premises correct?
\item[3.] Are all of the premises relevant?
\item[4.] Does the explanation introduce something new and useful, i.e., does more than just restate the conclusion in different words?
\item[5.] Which explanation do you prefer?
\end{des}
Answer options for questions 1-4 were yes/somewhat/no, and for question 5 were first/similar/second.
The ordering of explanations were scrambled so the annotators did not know which explanation was which,
and in fact they were unaware that the different explanations had been generated by different systems.
We collected annotations for 100 pairs of explanations for correct answers to 100 OBQA dev questions. \camera{The annotators were recruited from our institute, spanning a broad range of age (20-60) and experience.}

The results (Figure~\ref{user-judgements}) suggest several findings:

\begin{figure}
\centering
     \includegraphics[width=1\columnwidth]{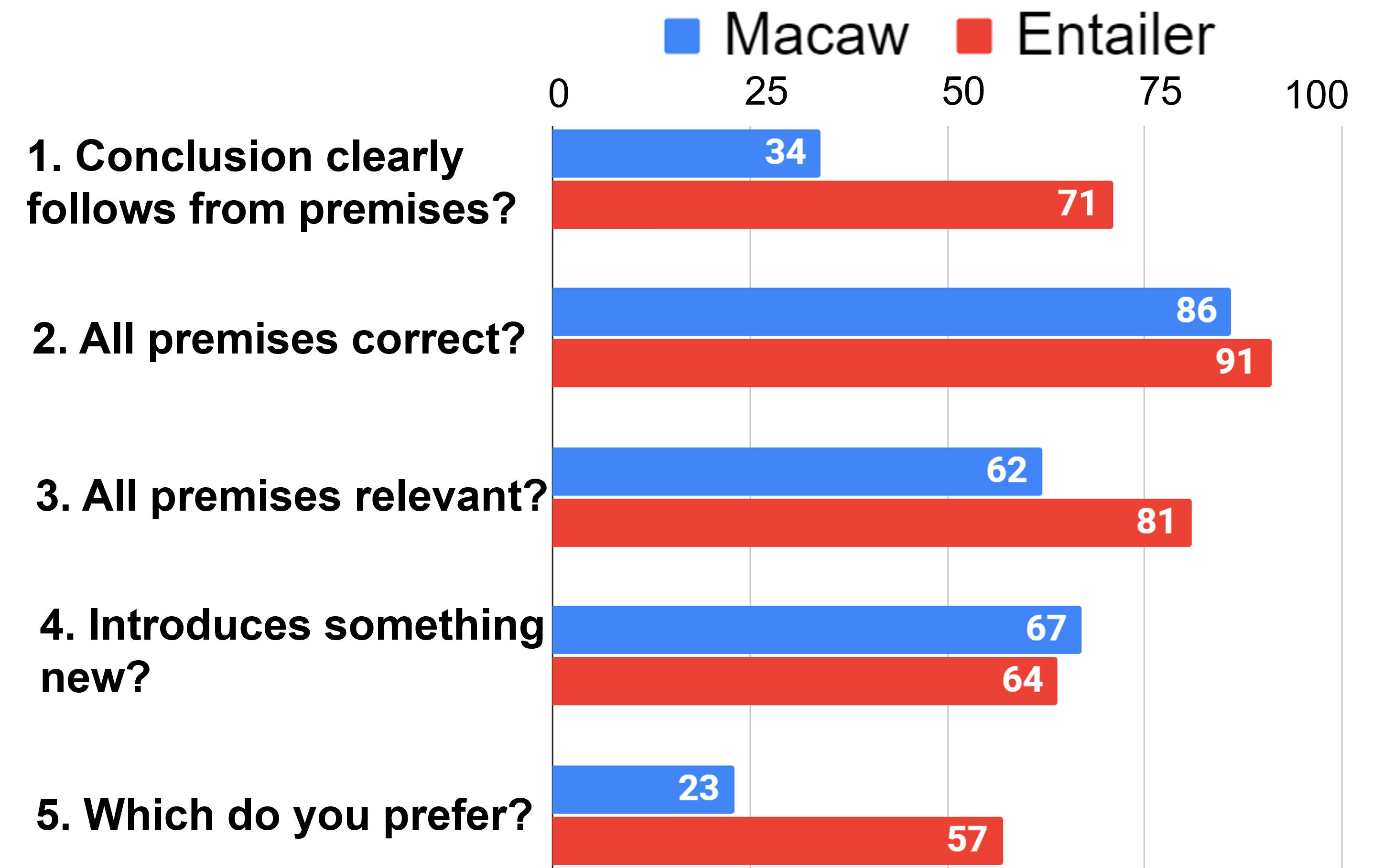}	   
\caption{Users' assessments of \EntailmentWriter{}'s proofs (red), compared to
Macaw's explanations (blue), showing percent of times annotators answered ``Yes''.
\EntailmentWriter{}'s conclusions were judged to ``clearly follow from the premises'' in over 70\% of the proofs
(first bar), substantially more than Macaw's explanations (34\%).
\label{user-judgements}}
\vspace{-3mm}
\end{figure}

\noindent
1. In absolute terms, \EntailmentWriter{}'s conclusions were judged to {\bf clearly follow from the premises in
the large majority (over 70\%)} of the explanations, and substantially more than Macaw's explanations (34\%).
This potentially contributes to system trustworthiness, where
understanding {\it how} evidence supports a conclusion is critical. \\
2. $\approx$90\% of \EntailmentWriter{}'s self-verified premises were judged correct by users. Of the remainder, virtually
all were labeled ``unsure'' (only 1 \EntailmentWriter{} fact was labeled not correct), indicating that the
there are {\bf few false beliefs} in proofs for correct answers. 
Rather, vague facts (e.g., ``Claws are used for cracking open shells'') make up the remainder. \\
3. {\bf \EntailmentWriter{}'s explanations were clearly preferred} (57\% to 23\%, last bar) over Macaw's.
In particular, \EntailmentWriter{}'s arrangement of facts into a tree discourages irrelevant information (bar \#3). \\
Finally we note \EntailmentWriter{}'s proofs are {\it faithful} (showing how an answer was derived)
and {\it truthful} (reflecting the system's own beliefs), while Macaw's explanations are post-hoc generated text.
These all suggest the value of entailment trees as trustable explanations of a system's behavior.

\subsection{Analysis}

\subsubsection{Failure Analysis \label{failure-analysis}}

If \EntailmentWriter{} answers a question incorrectly, either a model belief ({\bf belief error})
and/or the reasoning itself ({\bf reasoning error}) must necessarily be incorrect, unless the question itself is
malformed ({\bf dataset error}). To better understand these, we classified the 51/500 cases in OBQA
where \EntailmentWriter{} selected the wrong answer while the Direct answer was correct, and found: \\
1. {\bf belief errors (33\%)}, where an {\bf incorrect belief} led to a wrong answer, for example: \\
\smal
\fbox{\parbox{0.95\columnwidth}{
In a desert... plants grow closer together {\bf because} \\
1. A desert...contains a low amount of  water. \\
\red{2. As the amount of  water decreases, plants are forced to grow closer together to conserve water.}}}
\normalsize
Note that here the reasoning is correct but the second is premise is false. \\
2. {\bf dataset errors (20\%)}. In several cases, the question was ambiguous (e.g., does ``animal'' include
humans?) or more than one answer option was valid (OBQA is a crowdsourced dataset).  \\
3. {\bf reasoning errors (47\%):} \\
3a. {\bf Near-tautologies (20\%):} of the form ``$X$ {\bf because} 1. $X'$, 2. ...'',
where $X'$ is a near-repeat of the hypothesis. In such cases, the proof
offers little new information. \vspace{1mm} \\
3b. {\bf Basic reasoning errors (10\%):}, e.g.,
\noindent
\smal
\fbox{\parbox{0.95\columnwidth}{
\red{During ...fall, new leaves begin to grow} {\bf because} \\
1. During fall, old leaves begin to fall. \\
2. New leaves begin to grow.}}
\normalsize
Here entailment is simply invalid. \vspace{1mm} \\
3c. {\bf Incorrect abductive inferences (9\%):} of the form $A$ {\bf because} $(A\rightarrow{}B)$ and $B$.
While sometimes useful, these inferences are not sound and can produce incorrect conclusions, e.g., \\
\smal
\fbox{\parbox{0.95\columnwidth}{
\red{Cooking food requires a fire to be built} {\bf because}\\
1. Cooking food requires heating the food. \\
2. Fire can be used to heat things.}}	
\normalsize 
3d. {\bf Quantification and exceptions (8\%):} where both beliefs and reasoning
seem reasonable, but the specific case does not hold, e.g.,
\smal
\fbox{\parbox{0.95\columnwidth}{
\red{Seeds are often found inside a strawberry} {\bf because} \\
1. Seeds are often found inside fruits. \\
2. A strawberry is a kind of fruit.}}	
\normalsize

\eat{
belief: incorrect beliefs	33
dataset: ambiguous/bad question	20
reasoning: near-tautology	20
reasoning: bad entailment	10
reasoning: invalid abductive inference	9
reasoning: Quantification and exceptions	8
}

\begin{figure}
\centering
     \includegraphics[width=0.90\columnwidth]{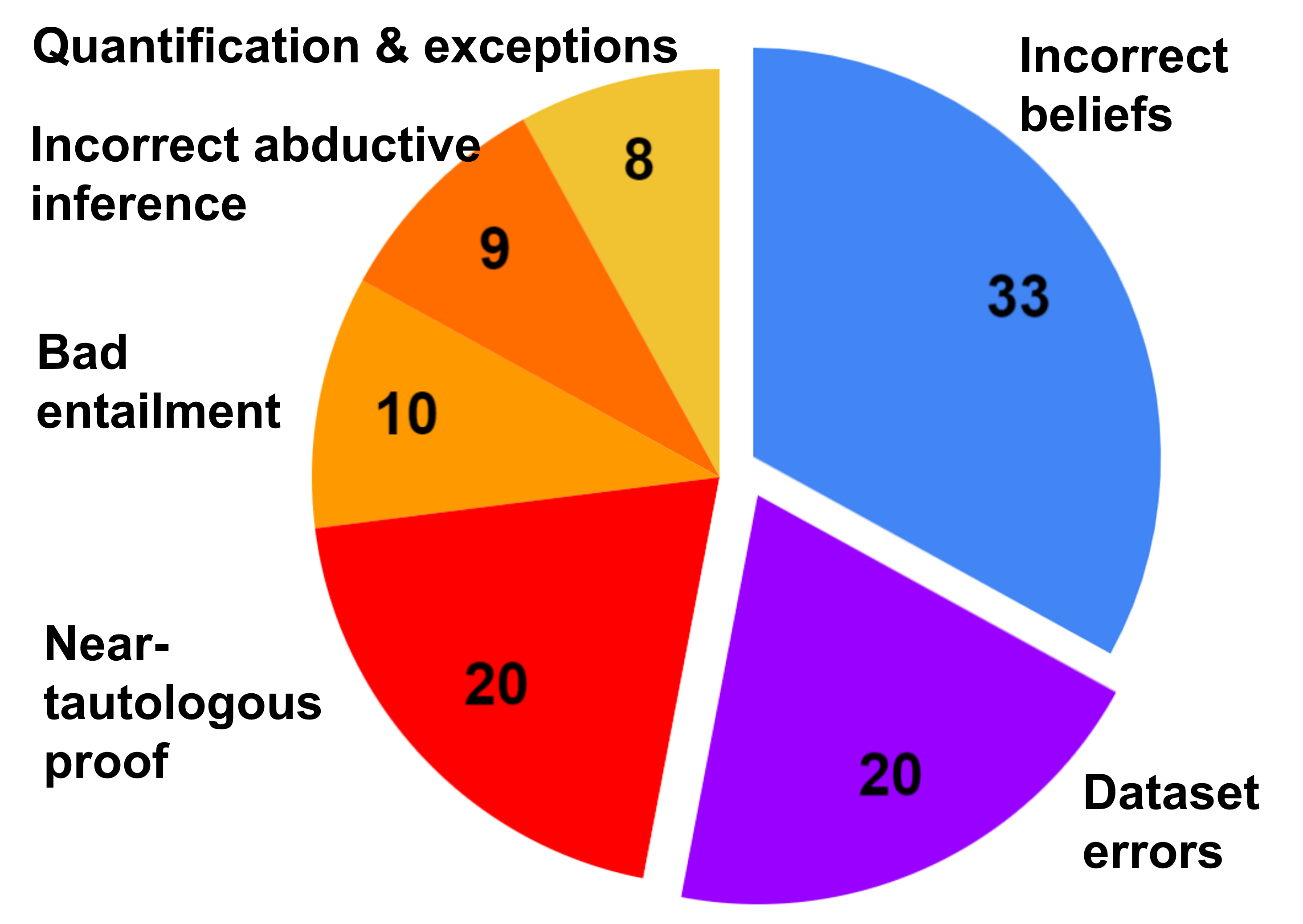}	   
\caption{Causes of \EntailmentWriter{}'s failures (\%) among belief (blue), reasoning (red shades), and dataset (purple). \label{failures}}
\vspace{-3mm}
\end{figure}


\eat{
  We also examined 20 of the 49/500 OBQA questions where \EntailmentWriter{}
produced the right answer, while Direct QA did not. Of these, the conclusions
clearly followed from the entailment proof in 18/20 (87\%) of the cases,
and in 17/20 \EntailmentWriter added new, correct, and useful information. These
figures are in line with the human annotation results in Figure~\ref{user-judgements}.
}

\subsubsection{When do proofs do better?}

At its best, \EntailmentWriter{} decomposes an uncertain hypothesis $H$
into premises $P$ which it is very confident about. For example, \EntailmentWriter{}
is unsure whether {\it A suit of armor conducts electricity}, but it
confidently believes the generated premises:\\
\smal
\fbox{\parbox{0.95\columnwidth}{
A suit of armor conducts electricty {\bf because} \\
1. A suit of armor is made of metal \\
2. Metals conduct electricity}}
\normalsize
Thus an uncertain conclusion is replaced by more certain premises, and we see
\EntailmentWriter{} performing particularly well for such questions. However, there
are questions that are largely ``fact lookup'', where a proof may be
less helpful. For example, the model is already very confident about the
hypothesis {\it Jupiter is the largest planet in the Solar System}; decomposing
this into {\it 1. The largest planet has the greatest mass}
plus {\it 2. Jupiter has the greatest mass in the solar system} has not obviously
made answering easier. In fact, \EntailmentWriter{}'s algorithm is specifically
designed to account for this, only growing a proof when the confidence in the
premises improves the confidence in $H$, thus tailoring its degree of reasoning
to the complexity of the problem at hand.

\section{Towards Teachable Systems}
\camera{
If a system can show how its answers systematically follows from its own beliefs,
articulating those beliefs in the process, then this opens a window into the model's
internal state for the user. As a result, new opportunities arise for identifying and
correcting the model's misunderstandings, so the system
gradually improves over time (sometimes referred to 
as explanatory interactive machine learning (XIL) \cite{xil}).
One vehicle for doing this is to use Entailer's (currently unused) context field C at runtime:
If the user asks a question, receives a wrong answer, and sees an incorrect belief in the proof,
they would provide the corrected belief, then re-ask the question with the corrected belief in the context. This encourages the model to use the corrected belief in its new answer and proof, rather than just repeat the same bad belief. Such overrides would then be stored and retrieved from a persistent memory to use for future questions also. A simple, hypothetical dialog illustrating this is shown in Figure~\ref{teachable}.
This is an exciting avenue made possible by this work, 
currently used by the TeachMe system\cite{teachme}.}

\begin{figure}[t]
\noindent
\fbox{\parbox{0.97\columnwidth}{\small{
\begin{des}
\item[{\bf U:}] \blue{Can a magnet attract a penny?} \\
{\it Entailer tries to prove H (A magnet attracts a penny) and the negation neg(H), and reports the best proof} 
\item[{\bf E:}] \blue{A magnet can attract a penny \bfit{because:} \\
1. A penny is made of copper \\
2. Copper is magnetic \\
Do you agree?} \\
{\it Here Entailer gives the wrong answer, arising from an incorrect belief \#2. The user offers a correction:} 
\item[{\bf U:}] \blue{No. Copper is not magnetic} \\
{\it System tries again, trying to prove H and neg(H) with context C = ``Copper is not magnetic''. C biases the model away from its prior mistake, effectively changing the model's belief. The new best proof is reported.}
\item[{\bf E:}] \blue{A magnet cannot attract a penny \bfit{because:} \\
1. A penny is made of copper. \\
2. Copper is not magnetic \\
Do you agree?} \\
  {\it The system now gets the right answer, having learned that copper is not magnetic from the
    user. C is added to memory for use in future questions, via information retrieva.}
\item[{\bf U:}] \blue{Yes}
\end{des}}}}
\caption{\label{teachable} 
  \camera{A hypothetical dialog illustrating how a user ({\bf U}) might identify
  and correct Entailer's ({\bf E}) incorrect beliefs through interaction.
  Here Entailer initially gets the wrong answer due to an incorrect
  model belief (``Copper is magnetic''). The user offers a correction,
  which is then provided as context when re-asking the question,
  effectively overriding the prior bad belief (the user has ``taught'' the model).
  By storing such corrections in a memory, such belief updates persist over time.}}
\end{figure}

\section{Conclusion}

Our goal is a system that can show how its answers systematically follow
from its own internal beliefs, and materialize those beliefs in the process.
\EntailmentWriter{} is, to our knowledge, the first system to demonstrate this capability,
achieved using an  ``over-generate and verify'' approach for each backward-chaining step.
Our evaluation suggests that \EntailmentWriter{}'s proof-based answer accuracy is
similar to the ``direct'' QA accuracy, with the additional benefit of providing a faithful,
truthful chain of reasoning showing how its answers follow from its
internal beliefs. The significance of this is that these chains
provide a window into the model's internal beliefs and their relationships,
allowing users to both verify a system's answer, or if the system
provides an incorrect answer, to identify the incorrect belief leading
to that error. This in turn offers new opportunities for a user to correct the
system's misunderstanding by overriding a faulty belief, e.g., by adding
a memory of user corrections/overrides \cite{prompt-editing}, or 
by editing the model parameters directly \cite{Mitchell2021FastME},
a step towards interactive systems that can both explain to,
and learn from, users over time \cite{teachme}. 
\camera{\EntailmentWriter{} data and models are available at \url{https://allenai.org/data/entailer}.}
We look forward to future developments
in these directions.





\section*{Limitations}

We have shown how a neural system can expose how its answers systematically 
follow from its own internal beliefs, providing a window into the model's
system of beliefs. While exciting, there are several limitations
with the current work and opportunities for the future.

First, the system is not perfect at generating coherent chains of reasoning,
sometimes producing entailments that are invalid or nearly tautologous
(Section~\ref{failure-analysis} and Figure~\ref{failures}). Improved proof generation and scoring
techniques would help address this.

Second, like others, we use textual entailment as the basic reasoning
operation, but the definition of entailment remains somewhat imprecise
(a valid entailment is one that ``a person would typically infer.''
\cite{Dagan2013RecognizingTE}), contributing to noise in the
model's training data. A more precise characterization of reasoning
validity would help in both generation and evaluation of reasoning
chains.

Third, we assume the model is generally consistent about its beliefs,
but in some cases the model may verify contradictory statements, making
it less clear what the model actually believes in such cases. We
currently do not handle such situations. Use of a global notion of belief
(rather than per question) would be a valuable avenue to explore, e.g., \cite{beliefbank}.

Fourth, as a practical matter, recursive construction of proofs
is computationally expensive ($\approx$360 seconds/question for up to depth-3
proofs for 4-way multiple-choice,  Appendix~A.2).
Improvements to the search algorithm would allow faster experimentation,
and also help deploy the model in a practical setting.

Finally, we have speculated that showing users faithful, truthful
chains of reasoning might be a basis for a conversational system,
where users could correct and teach the system in cases where it
was wrong. However, this is currently just a conjecture -
futures explorations into how this might be realized would be valuable.

\eat{
Limitations

Paper 1
 - Bad proofs slip through the entailment verifier
 - The scoring is imperfect, even after training
 - We are getting comparable but not superior score
 - Entailment proofs can be vacuous
 - We've used a simple notion of belief: if the KB is inconsistent, it gets more fuzzy
 - Algorithm is slow

Future work 
 - Algorithm improvements: bad scoring, vacuous proofs
 - Beliefs and consistency of beliefs not accounted for; belief management
 - Runtime Speed
 - Interaction with the system would be nice
}

\camera{
\section*{Ethics Statement}
Like any other large-scale
language model, despite the best intentions, there is a risk of our model producing biased or offensive statements as part of its explanations. We release our models for research purposes only.}

\section*{Acknowledgements}
This research was made possible, in part, by funding from Open Philanthropy.
We also thank Google for providing the TPUs for conducting experiments.
\bibliography{references}
\bibliographystyle{acl_natbib}

\clearpage

\appendix

\onecolumn


\section*{Appendix A. \EntailmentWriter{}'s Backward Chaining Algorithm}
    
\algrenewcommand\algorithmicindent{1.0em}%
\algrenewcommand\algorithmiccomment[1]{\hfill {\it // #1}}

\begin{algorithm*}[ht!]
\begin{algorithmic}[1]
\small{
\Procedure{prove}{$H$} \ra{} score $s(H)$ \& proof $tree(H)$:
\State i. {\bf find} \underline{direct} score $s_{d}(H)$ \& confidence $c_{d}(H)$:
\State \gapxx         direct score $s_{d}(H)$ = {\bf call} model $H$\ra{}$S_d$
\State \gapxx         direct confidence
$c_{d}(H)$ = $max(s_d(H),(1-s_d(H)))$ \Comment{confidence of predicted label}
\State ii. {\bf find} \underline{reasoned} score $s_{r}(H)$, confidence $c_r(H)$, \& subtrees $tree(p_i)$:
\State \gapxx       {\bf call} \textproc{1step}(H) \ra $P$ to find best premises $P$ that entail $H$
\State \gapxx       {\bf find} $s_e(P\vdash{}H)$ via model $PH\rightarrow{}S_{e}$           \Comment{score the entailment}
\State \gapxx       {\bf if} $s_e(P\vdash{}H) > c_d(H)$ {\bf or} $H$=$H_0^{\dagger}$ \Comment{if reasoning conf $c_r(H)$ might beat direct conf $c_d(H)$ (lines 13,11), backchain}
\State \gapxxx       {\bf then forall} $p_i \in P$ {\bf do} 
\State \gapxxxxxxxx        {\bf call} \textproc{prove}($p_i$) to find $s(p_i),tree(p_{i})$  \Comment{get scores and subtrees for each $p_i$}
\State \gapxxxxx 	    reasoned score $s_r(H) = (\Pi_{i} s(p_i)).s_e(P\vdash{}H)$                  \Comment{overall score for reasoned answer}
\State \gapxxx      {\bf else} $s_r(H) = 0$
\State \gapxxx        reasoning confidence $c_{r}(H) = s_{r}(H)$ \Comment{reason confidence = score}
\State iii. {\bf return} direct or reasoned answer (pick most confident):
\State \gapxx        {\bf if} $c_{r}(H) > c_{d}(H)$ {\bf or} $H$=$H_0^{\dagger}$ \Comment{if reasoning confidence is higher (or top level)}
\State \gapxx        {\bf then return} $s(H)$=$s_{r}(H)$ \& $tree(H)$=($\{tree(p_{i})\} \vdash{} H$) \Comment{reasoned score + subtrees}
\State \gapxx        \hspace*{0.5mm} {\bf else return} $s(H)$=$s_{d}(H)$ \& $tree(H)$=$H$ \Comment{direct score + H as terminal node (the subtrees are discarded)}
\EndProcedure
\State
\Procedure{1step}{$H$} $\rightarrow$ premises $P$  \Comment{Find premises $P$ that together entail $H$}
\State {\bf repeat} $k$ times: \Comment{over-generate with nucleus sampling}
\State \gapx        a. find premises $P$ via model $H\rightarrow{}P$ \Comment{find premises}
\State \gapx        b. find $\{s_d(p_i \in P)\}$ via model $H(=p_{i})\rightarrow{}S_{d}$ \Comment{score those premises}
\State \gapx        c. find $s_e(P\vdash{}H)$ via model $PH\rightarrow{}S_{e}$           \Comment{score the entailment itself}
\State \gapx        d. discard if any of $\{s_d(p_i)\},s_e(P\vdash{}H) < 0.5$       \Comment{now filter}
\State \gapx        e. $s_{r\textnormal{-}1deep}(H) = (\Pi_{i} s_d(p_i)).s_e(P\vdash{}H)$         \Comment{Final score for the 1-deep proof}
\State {\bf select} premises $P$ with highest $s_{r\textnormal{-}1deep}(H)$  \Comment{Pick the best of the $k$ proofs}
\EndProcedure
}
\end{algorithmic}
\footnotesize{
$^{\dagger}$We insist on at least a 1-deep tree by, for the top-level hypothesis $H$=$H_0$, ensuring that lines 8 and 15 succeed.}

\caption{\EntailmentWriter{}'s backchaining algorithm for searching for the best proof $tree(H)$ with score $s(H)$ for a hypothesis $H$. \label{backward-chaining}}
\end{algorithm*}

\begin{multicols}{2}

\subsection*{A.1 Generating One Backward-Chaining Step}

The procedure to find a 1-step inference is called \textsc{1step}($H$) in Algorithm~\ref{backward-chaining} (line 19).
Given a hypothesis $H$, we use the angle $H \rightarrow P$ to over-generate a set of $k$ alternative backward-chaining steps $P \vdash H$ using
nucleus sampling (line 21). We then check that the model believes all the premises $p_i \in P$ using the $H (=p_i) \rightarrow S_{d}$ angle,
and that it believes the inference step $P \vdash H$ itself is valid (independent of whether the $p_i$ are true or not) using the $PH \rightarrow S_{e}$ angle (line 24).
The overall score, denoting how well the 1-step proof supports the hypothesis, combines the premise end entailment scores as follows (line 25):
\begin{myquote}
\vspace{1mm}
\sgap $s_{r\textnormal{-}1deep}(H) = (\Pi_{i} s_d(p_i)).s_e(P\vdash{}H)$
\end{myquote}
\vspace{1mm}
The highest-scoring proof $P\vdash H$ is returned.

\subsection*{A.2 Backward Chaining}

Procedure \textsc{prove}($H$) in Algorithm~\ref{backward-chaining}
generalizes this to multi-step entailments by
recursively generating support for each premise $p_{i}$ in $P$ (line 10).
The stopping condition is when the model's direct confidence\footnote{
As the direct score $s_d(p_i)$ in $p_i$ ranges from 0 (definitely false) to 1 (definitely true),
we define the corresponding confidence to with respect to the predicted label, i.e., $c_d(p_i) = 1-s_{d}(p_i)$ when $s_{d}(p_i)<0.5$, otherwise $c_d(p_i) = s_{d}(p_i)$.
In contrast, the proof score ranges from 0 (no proof) to 1 (perfect proof), hence the
proof confidence is simply the proof score, $c_r(p_i) = s_r(p_i)$. Note that unlike $s_d(p_i)$, $s_r(p_i)=0$ means no proof, not false.}
in $p_{i}$ is greater than the proof-derived confidence in $p_{i}$,
i.e., $c_d(p_i) > c_r(p_i)$. When this condition 
is met, the subtree $P' \vdash p_{i}$ for $p_{i}$ is discarded and
$p_{i}$ becomes a leaf node of the tree (line 17). We also impose a maximum
depth $d$ on the tree.

This whole procedure is repeated for each candidate answer hypothesis (Section~\ref{qa2d}).
Finally the system selects the answer corresponding to the hypothesis with the top-scoring proof $s(H)$. 

On our datasets, the average runtime per question (4-way multiple-choice) is $\approx$80 seconds (depth 1 proofs, sample size $k$=6) or $\approx$360 seconds (up to depth 3 proofs, sample size $k$=6) on a 48GB GPU, due to the large number of candidate inference steps generated during the search.

\end{multicols}

\onecolumn

\section*{Appendix B. Crowdsourcing Instructions for Verifying Entailments (Section~\ref{crowdsourced-data})}

\includegraphics[width=0.9\columnwidth]{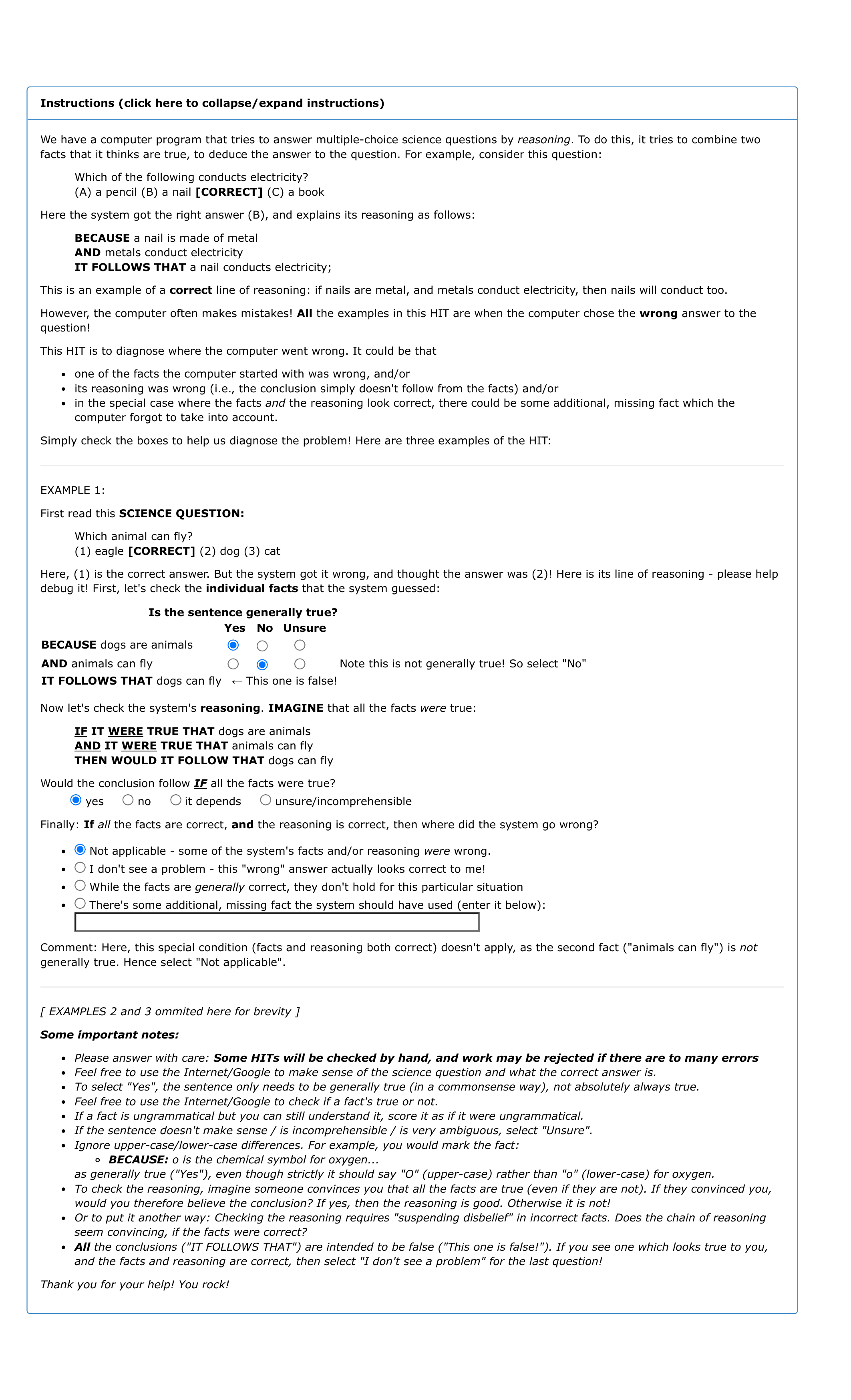}


\twocolumn

\section*{Appendix C: Model Training}

\subsection*{C.1 Dataset Preparation}

Here we describe in detail how \EntailmentWriter{}'s training data is assembled.

\begin{enumerate}
\item We start with the training partition of the EntailmentBank dataset, containining 1313 multiple choice questions
each with an entailment tree for their correct answer choice. 

\item We convert the question + each answer option to a hypothesis using four different, alternative methods:
  \begin{ite}
  \item \EntailmentWriter{}'s current QA2D model, a reconstruction of that by \cite{qa2d} (Section~\ref{qa2d}).
  \item An in-house, rule-based QA2D utility
  \item An earlier version of \EntailmentWriter{}'s QA2D model.
  \item The original hand-written hypothesis supplied in the EntailmentBank dataset.
    \end{ite}
If a source generates the same hypothesis for two different answer choices, we choose not to trust it for that question and discard it.

\item We ``shred'' the EntailmentBank entailment trees into individual 1-step P $\vdash$ H entailments,
  producing 4175 1-step (valid) entailments using $\approx$9000 true premises.

\item The crowdsourced P $\vdash$ H instances, with annotations on whether premises are true and entailment is true, are also added (Section~\ref{crowdsourced-data}).

\item Each EntailmentBank proof also comes with a set of associated relevant facts (sentences), only some of which are used in the entailment proofs.
  We use these to create a "relevant context" containing these sentences, sorted into two buckets:
  \begin{des}
    \item[{\bf high-relevance:}] sentences actually used in the proof of the correct answer
    \item[{\bf medium-relevance:}] the remaining sentences
  \end{des}
This context is the same across all answer options at this point. 


\item For every hypothesis and premise appearing in this dataset, we run a simple BM25 IR search against a larger science text corpus (about 1.5 million sentences from a science-filtered Wikipedia) to obtain noisier sentences to use as a {\bf low-relevance} context.

\item For the final training dataset, we create 4 different contexts, using 4 sampling strategies. First we create a "full" context of sentences sorted into three buckets,
  with added noise:
  \begin{des}
  \item[{\bf high:}] high-relevance facts + 10\% chance of a random sentence from medium/low
  \item[{\bf medium:}] medium-relevance facts + 20\% chance of random sentences from high/low
  \item[{\bf low:}] 5 low-relevance facts + 20\% chance of random sentence from high/medium
  \end{des}
  Then, we use four sampling strategies to create actual contexts for the \EntailmentWriter{} training data.
  Each strategy is defined by the per-sentence chance of a sentence coming from one of the high/medium/low buckets.
  For example, a context sampled with 0.2/0.4/0.6 means that 20\% of its sentences came from high, 40\% from medium, an 60\% from low. If no sentence is selectable (e.g., a bucket is exhausted), we use one from low. The final context is syntactically expressed as ``[HIGH] {\it <high sentences>} [MEDIUM] {\it <medium sentences>} [LOW] {\it <low sentences>}''. The 4 sampling strategies are:
  \begin{des}
   \item[{\bf 1/1/1}]  (full context, all available sentences)
   \item[{\bf 0.5/0.5/0.5}]  (half of sentences from each category)
   \item[{\bf 0.2/0.4/0.6}]  (more emphasis on lower categories)
   \item[{\bf 0/0/1}]  (only low sentences)
  \end{des}
For the training set we store these as a list of contexts, to be sampled at random when generating the training instances. We do the same for hypotheses, so these are also sampled at random.

\item The model is trained across the following angles, each sampled equally:
\{H\ra{}P, H\ra{}V, HP\ra{}I, QAH\ra{}P, QAH\ra{}V, QAHP\ra{}I, HC\ra{}P, HC\ra{}V, HPC\ra{}I, QAHC\ra{}P, QAHC\ra{}V, QAHPC\ra{}I\}

This is just the core angles H\ra{}P, H\ra{}V, HP\ra{}I, with optional QA and/or C added. The individual premise verification (leaves) uses the H\ra{}V angle (where H is now a premise), for those angles we don't have associated proofs P so we limit to the angles: H\ra{}V, QAH\ra{}V, HC\ra{}V, QAHC\ra{}V

The full dataset is provided in the supplementary material, and will be released publically.

\end{enumerate}

\subsection*{C.2 Model Details}

We train a T5-11B multi-angle model, \EntailmentWriter{}, following the multi-task setup similar to \cite{macaw}
for the three functionalities described in Table~\ref{angles}.
We used the default hyperparameters (including the Adafactor optimizer) in the T5 library,
fine-tune the model for 20K steps with batch size of 8, selecting the checkpoint with highest validation score (usually the final step).

At inference time, \EntailmentWriter{}'s primary function is to generate candidate entailments for each answer choice for a given question, which are then scored by the same model using its two verification angles. To generate multiple explanations for a single input (answer hypothesis, IR retrieved context), in addition to the greedy decoding we 
use nucleus sampling \cite{nucleus-sampling}. For the experiments in this paper we set ``temperature= 2.0'', and ``top\_p= 0.95''.


\onecolumn

\clearpage

\section*{Appendix D. Examples of Macaw Explanations and \EntailmentWriter{} Proofs}

A random selection of explanations, both good and bad, from the two systems (Section~\ref{human-judgements}).

\vspace{2mm}

\includegraphics[width=0.95\textwidth]{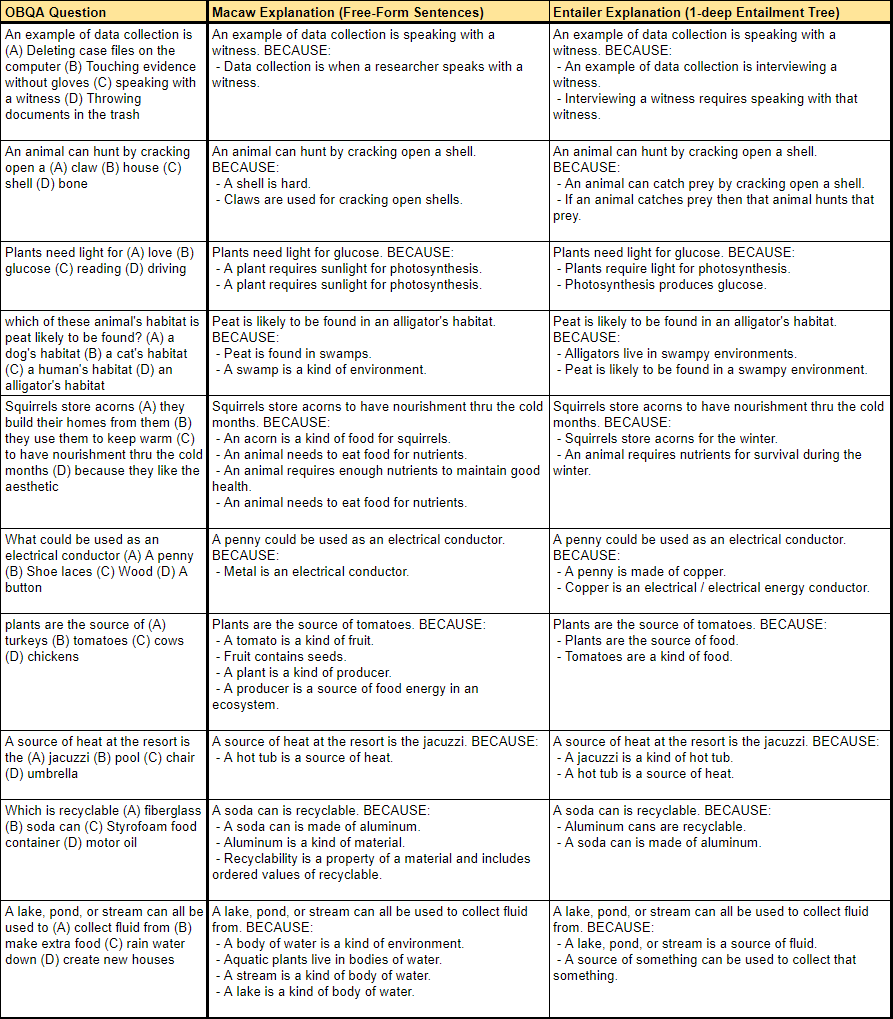}	   

\end{document}